\documentclass[runningheads]{llncs}

\usepackage{eccv}

\usepackage{makecell}
\usepackage{tabularx}
\usepackage{multirow}
\usepackage{pifont}
\usepackage{array}
\usepackage{adjustbox}
\usepackage{xcolor}
\usepackage{amsmath}
\usepackage{url}            
\usepackage{booktabs}       
\usepackage{amsfonts}       
\usepackage{nicefrac}       
\usepackage{microtype}      
\usepackage{xcolor}         
\usepackage{graphicx}
\usepackage{amsmath}
\usepackage{bm}
\usepackage{multirow}
\usepackage{tabu}
\usepackage{siunitx}
\usepackage{adjustbox}
\usepackage{subcaption}
\usepackage{siunitx}
\usepackage{xcolor}
\usepackage{colortbl}
\usepackage{color}
\usepackage{pifont}
\usepackage{colortbl}
\definecolor{Gray}{gray}{0.95}
\definecolor{Cyan}{rgb}{0.88,1,1}
\usepackage{wrapfig}
\usepackage{algorithm}
\usepackage{algpseudocode}
\usepackage{tabu}           
\usepackage{multirow}
\usepackage{booktabs}
\newcommand{\paragrapha}[2][4pt]{\vspace{#1}\noindent\textbf{#2}}
\newcommand{\bfsection}[1]{\noindent\textbf{#1.}}

\newcommand\cb[1]{\color{blue} #1}
\newcolumntype{x}[1]{>{\centering\arraybackslash}p{#1pt}}
\newlength\savewidth\newcommand\shline{\noalign{\global\savewidth\arrayrulewidth
  \global\arrayrulewidth 1pt}\hline\noalign{\global\arrayrulewidth\savewidth}}

\newcommand{\tablestyle}[2]{\setlength{\tabcolsep}{#1}\renewcommand{\arraystretch}{#2}\centering\footnotesize}
\usepackage{pifont}
\usepackage{eccvabbrv}

\usepackage{graphicx}
\usepackage{booktabs}

\usepackage[accsupp]{axessibility}  

\usepackage{hyperref}

\usepackage{orcidlink}

\makeatletter
\def\thanks#1{\protected@xdef\@thanks{\@thanks
        \protect\footnotetext{#1}}}
\makeatother

\begin{document}

\title{Efficient Inference of Vision Instruction-Following Models with Elastic Cache} 

\titlerunning{Elastic Cache}

\author{Zuyan Liu\inst{1}\thanks{\textsuperscript{\dag}Corresponding author.}\orcidlink{0009-0002-6943-3085} \and
Benlin Liu\inst{2} \and
Jiahui Wang\inst{1} \and
Yuhao Dong\inst{1,5} \and \\
Guangyi Chen\inst{3,4}\orcidlink{0000-0001-7542-5378} \and
Yongming Rao\inst{1,5}\orcidlink{0000-0003-3952-8753} \and
Ranjay Krishna\inst{2,6} \and
Jiwen Lu\inst{1}$^{\dagger}$\orcidlink{0000-0002-6121-5529}
}

\authorrunning{Z.~Liu et al.}

\institute{Tsinghua University, \and
University of Washington, \and
Carnegie Mellon University, \\ \and
Mohamed bin Zayed University of Artificial Intelligence, \\ \and
Tencent, and \and 
Allen Institute for AI
}

\maketitle

\begin{abstract}
In the field of instruction-following large vision-language models (LVLMs), the efficient deployment of these models faces challenges, notably due to the high memory demands of their key-value (KV) caches. Conventional cache management strategies for LLMs focus on cache eviction, which often fails to address the specific needs of multimodal instruction-following models. Recognizing this gap, in this paper, we introduce Elastic Cache, a novel approach that benefits from applying distinct acceleration methods for instruction encoding and output generation stages. We investigate the metrics of importance in different stages and propose an ‘importance-driven cache merging’ strategy to prune redundancy caches. Instead of discarding less important caches, our strategy identifies important key/value vectors as anchor points. Surrounding less important caches are then merged with these anchors, enhancing the preservation of contextual information in the KV caches while yielding an arbitrary acceleration ratio. For instruction encoding, we utilize the frequency to evaluate the importance of caches. Regarding output generation, we prioritize tokens based on their ‘distance’ with an offset, by which both the initial and most recent tokens are retained. Results on a range of LVLMs demonstrate that Elastic Cache not only boosts efficiency but also notably outperforms existing pruning methods in language generation across various tasks. Code is available at \url{https://github.com/liuzuyan/ElasticCache}
\end{abstract}

\section{Introduction}
\label{sec:intro}

ChatGPT~\cite{brown2020language,OpenAI_GPT4_2023} has rapidly gained popularity for its coherent and fluent responses.
Its effectiveness stems from an instruction-following LLM~\cite{ouyang2022training, chiang2023vicuna}, which handles diverse tasks based on input instructions. 
The model can also integrate visual inputs, as seen in GPT-4V~\cite{OpenAI_GPT4V} and LLaVA~\cite{liu2023visual}, expanding its applications, including multimodal chatbots.

However, multimodal instruction-following models' high computational and memory demands pose a challenge. These demands are critical in dialogue systems, where real-time responsiveness is essential for user experience. 
Therefore, the need to enhance the efficiency of these models becomes increasingly evident, particularly when generating lengthy outputs, given that the complexity of this task is compounded by the quadratic computational demands of attention modules in transformers~\cite{vaswani2017attention}.
To address this, a KV Cache mechanism is used in generative inference, storing and reusing key/value vectors for prompt and output tokens to reduce redundant computations.

\begin{figure}[tb]
  \centering
  \includegraphics[width=\linewidth]{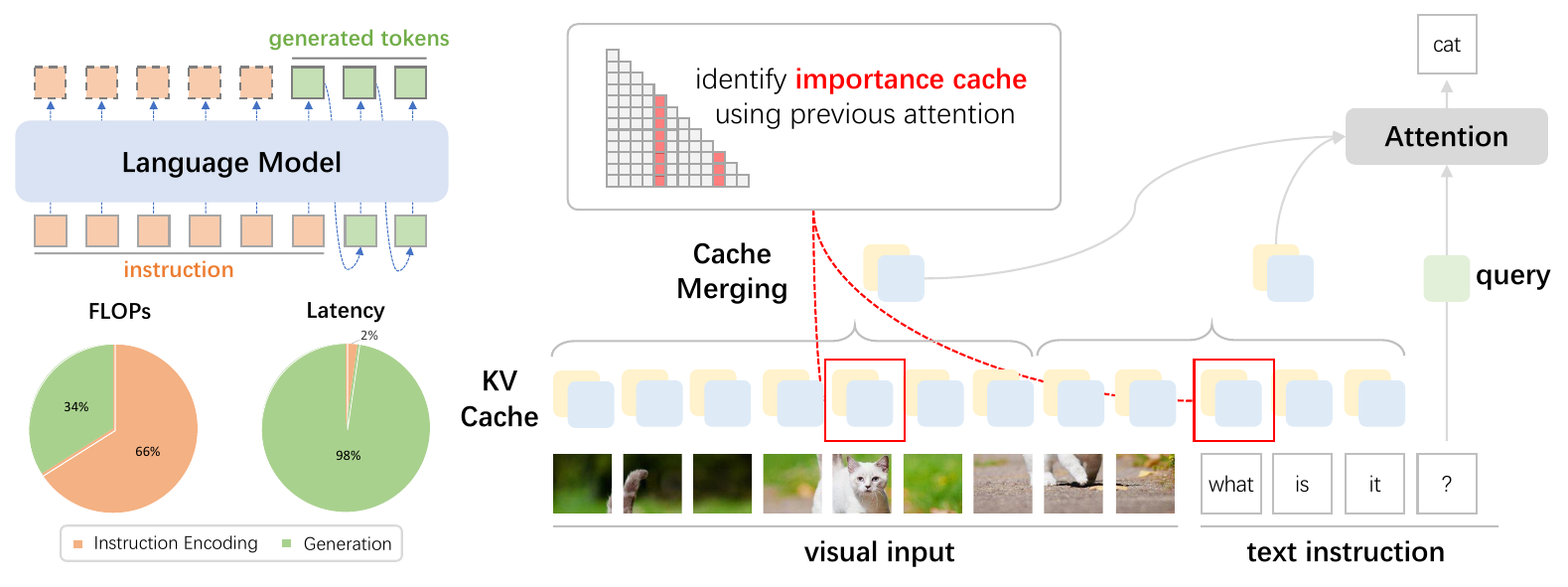}
  \caption{\textbf{The main idea of \emph{Elastic Cache}. }Instruction encoding accounts for most of the theoretical computation cost, while the actual latency is negligible (here we use generating 512 tokens based on a 1024-token instruction as an example). This underscores that it's not just model weights but also the KV cache used in output generation that can become a significant bottleneck. We propose \emph{Elastic Cache} through a Cache Merging based on the importance scores of instruction tokens, complemented by a fixed-point elimination strategy in the output generation phase. Our designs yield significant inference acceleration while maintaining generation quality.}
  \label{fig:intro}
\end{figure}

While effective, this widely used space-for-time strategy in KV cache management often leads to substantial GPU memory usage, sometimes exceeding the memory required for model weights. This can limit batch sizes and affect inference throughputs. 
Besides solutions like offloading KV cache to the CPU~\cite{aminabadi2022deepspeed} or cache quantization~\cite{sheng2023high} exist, recent studies~\cite{zhang2023h,liu2023scissorhands} explore pruning key/value vectors in the KV cache to reduce memory usage while maintaining language modeling performance. 
Such methods lower memory demands and improve computational efficiency, as they involve fewer vectors in attention calculations. 
Techniques like $\text{H}_2\text{O}$~\cite{zhang2023h} and Scissorhands~\cite{liu2023scissorhands} utilize smaller KV caches. When input sequences surpass the reduced cache size, these methods prune less important tokens based on attention scores, enhancing both memory and computational efficiency.

These existing methods have two main shortcomings. 
Firstly, their time/memory efficiency can be enhanced. They currently only improve computational and storage efficiency when the sequence length surpasses the KV cache's maximum capacity, with the acceleration ratio largely tied to this capacity. Our goal is to boost efficiency for any sequence length, independent of the cache size, thus improving efficiency even with a large cache for performance. 
Secondly, as stated in \cref{fig:main_result} and \cref{tab:generation} these methods don't specifically address maintaining the model's capability to generate long, coherent outputs that follow instructions, particularly for multimodal instructions.

To boost time and memory efficiency in multimodal instruction-following models for any sequence length, independent of cache budget, and maintain their multimodal instruction-following ability, we introduce a new KV cache management technique, \emph{Elastic Cache}. 
The essence of our method is the use of distinct sparsification strategies during instruction encoding and output generation phases. This distinction allows our model to better adhere to multimodal instructions with a compressed KV cache, surpassing previous methods that uniformly pruned KV vectors in both phases. Additionally, our approach enhances efficiency for sequences of all lengths by reducing token storage in the cache earlier during instruction encoding rather than waiting for the cache to fill.

Specifically, during the instruction encoding phase, we globally apply sparsification to prune key/value vectors generated from all concatenated model inputs, including system prompts, user instructions, and chat history. Unlike previous cache compression methods that used eviction strategies to reduce the number of key/value vectors stored in the KV cache, we introduce a novel parameter-free cache merging approach to more effectively preserve context in a compact KV cache. 
In detail, we determine the importance of all key/value vectors from the instruction encoding phase and utilize the most crucial ones as anchor points. Subsequently, we merge all key/value vectors in the entire instruction sequence with their nearest anchor point. A layer-wise merging policy is adopted, with all attention heads in the same layer sharing anchor point positions, although the values of anchor points at the same position may vary across different heads. We can achieve an arbitrary acceleration ratio by controlling the proportion of anchors in the cache merging process.
During the output generation phase, we dynamically manage the KV cache by adding new and removing older key/value vectors as tokens are generated. 
Unlike $\text{H}_2\text{O}$~\cite{zhang2023h}, our method employs a fixed-point elimination strategy at a tunable truncation point, balancing the cache between initial guidance and new content. This approach, akin to StreamingLLM~\cite{xiao2023efficient}, differs by retaining a length of initial vectors nearly equal to the input instruction length, better-preserving context to follow instructions.

Remarkably, our method is completely training-free, requiring no additional fine-tuning, and can be applied plug-and-play to any multimodal instruction-following model. This significantly saves the expenses associated with training extra-large models. Additionally, our cache update strategy incurs only negligible computational overhead during inference.

In our experimental analysis, we implemented the \emph{Elastic Cache} method in visual instruction-following tasks, employing Perplexity (PPL) and ROUGE as our primary evaluation metrics to assess instruction-following capabilities. The results demonstrate \emph{Elastic Cache}'s ability to significantly accelerate processing speeds without compromising on the quality of instruction following or generating lengthy outputs. It consistently outperforms both distance-based and frequency-based cache strategies across various models and datasets. Notably, with the KV Cache Budget set to 0.2, \emph{Elastic Cache} achieves a 78\% increase in actual speed. This marked improvement in processing efficiency, along with maintained or even enhanced predictive accuracy, highlights the practicality and effectiveness of \emph{Elastic Cache} in real-world scenarios, where time efficiency is as important as performance.

\section{Related Work}
\label{sec:formatting}
\noindent\textbf{Vision Instruction-Following Model}
After pre-training on trillions of tokens by predicting the next token, decoder-only large language models~\cite{brown2020language,touvron2023llama,xwin-lm} demonstrate an astonishing understanding of language. To better enable language models to follow instructions and generalize zero-shot to new tasks~\cite{ouyang2022training,wei2021finetuned,yang2024learning,liu2024chain,yang2023octopus,alpaca}, these models are further instruction tuned on human-annotated 'instructional' data, which includes language instructional commands and desired outcomes. Recently, how to enable these blind language models to follow instructions containing visual signals has become a hot research topic. Compared to naively using image caption models for prompting to integrate visual signals~\cite{shen2023hugginggpt,yang2023mm}, works like LLaVA~\cite{liu2023visual}, MiniGPT-4~\cite{zhu2023minigpt}, Multimodal-GPT~\cite{Gong2023MultiModal}, InstructBLIP~\cite{instructblip} use linear projection or perceivers~\cite{jaegle2021perceiver} to integrate visual representations into the input of LLMs directly. This allows the model to follow multimodal instructions better. Our work focuses on improving the inference efficiency of multimodal instruction-following model~\cite{liu2023visual,Bai2023QwenVLAF}, especially in generating lengthy responses, which is particularly important for applications like multimodal chatbots.

\noindent\textbf{Model Compression and KV Cache Management.} 
Improving inference efficiency through model compression has always been a crucial topic for the practical application of deep learning models~\cite{han2015deep}. Like other models, LLM compression methods typically include quantization~\cite{lin2023awq,xiao2023smoothquant}, pruning~\cite{sun2023simple, ma2023llm}, distillation~\cite{hsieh2023distilling}, etc. These methods are usually orthogonal and can be combined. However, past LLM compression methods have primarily focused on the computational overhead of model weights. Using a KV cache to speed up inference also results in a significant additional demand for memory during the inference stage, thereby affecting the throughput of the inference system. To address this, besides using quantization~\cite{sheng2023high} or offloading~\cite{aminabadi2022deepspeed}, more research is focusing on reducing the number of KV vectors stored in the cache. Gist tokens~\cite{mu2023learning} learn to compress input instruction tokens before storing them in the KV cache, but this method requires additional training, and its cache update strategy, which includes extra parameter computations, is costly. StreamingLLM~\cite{xiao2023efficient} focuses on processing long sequences that exceed the cache's capacity limit, so it only can accelerate sequences with tens of thousands of tokens. $\text{H}_2\text{O}$~\cite{zhang2023h} and Scissorhands~\cite{liu2023scissorhands} achieve greater efficiency improvements by lowering the cache capacity limit, but they also can't accelerate sequences of arbitrary lengths. Our method is committed to accelerating sequences of any length while maintaining the capability to follow multimodal instructions and generate lengthy responses. Moreover, we employ a novel cache merging method for cache updates instead of the cache eviction approach used in previous works.

\section{Method}

\subsection{Preliminaries: Instruction-Following Models}

Instruction-following large vision-language models (LVLMs) are a cornerstone in natural language processing, especially when it comes to aligning with specific user intents and tasks. These models are invaluable in textual contexts for their adeptness at interpreting and executing instructions and demonstrate comparable value in multimodal settings. The core process of their inference involves two main stages: instruction encoding and output generation.

\paragrapha{Instruction Encoding. }  This is the foundational step in the operation of instruction-following models. During this phase, the model receives and interprets the system prompts, user-provided instructions, and chat history, which may include multimodal information.
For autoregressive transformers, this involves processing the instruction inputs and encoding them into a series of tokens. Each token is then transformed into key and value vectors, which are vital for capturing the contextual nuances of the instruction. These vectors are stored in the KV cache, ensuring that the model has immediate access to the contextual information needed for generating accurate responses.

\paragrapha{Output Generation. } After the instruction has been encoded, the instruction-following model proceeds to the output generation phase. Here, the model incrementally builds its response, token by token. At each step of this process, the model references the KV cache to ensure that each new token generated is contextually aligned with the preceding ones. This incremental approach allows the model to maintain a coherent narrative or logical thread in its responses while staying true to the user's initial instructions.

The operation of the instruction-following model is challenged by the high memory demands of their KV caches. These caches store all encoded tokens' key and value vectors, which are crucial for contextual understanding during instruction encoding and output generation. As interactions progress, the cache size grows linearly, significantly straining memory resources in complex or lengthy tasks. 
To address this, developing effective cache eviction strategies is crucial for balancing the efficiency and accuracy in generating responses of instruction-following models.

\subsection{Instruction following with Cache Merging}
In this section, we introduce the \emph{importance-driven cache merging} policy, serving as our core proposal for cache management. 
Its intuitive principle is finding the key/value vectors in the cache as the anchor point to aggregate the surrounding contextual information. 
Specifically, we mainly discuss two questions: 1) How do we measure the importance, and 2) Why do we employ the proposed cache merging instead of conventional cache eviction. 

\bfsection{Importance Metrics}
We propose to measure the importance of the key/value vectors in the KV cache using their statistical data, motivated by the stationary nature of the statistical information throughout the inference process. 
This stability suggests that historical statistical data can be effectively used to forecast future cache requirements. By leveraging past usage patterns within the KV caches, we can develop policies to predict which vectors in the cache will be most pertinent in upcoming inference tasks, thereby optimizing cache management and resource allocation.

Building on our initial findings, we conducted a thorough investigation to identify the most crucial statistic for evaluating the importance of KV-caches. Our experiments span an array of statistics, encompassing both population types—identical, head-wise, and layer-wise populations—and statistical variables, including summation, moving average, maximum value, mean, and others. The comparative analysis of these different metrics was meticulously detailed in the ablation studies presented in \cref{sec:ablation}. These investigations revealed that the layer-wise sum of attention scores yielded the most effective performance.

\bfsection{Cache Eviction v.s. Cache Merging}
This policy should be strategically designed to maintain the integrity and efficiency of the KV cache system, ensuring that the most valuable cached vectors are retained and readily accessible for future inference processes while less critical caches are efficiently removed to optimize resource allocation and system performance.

Conventional cache management strategies, like those in StreamingLLM~\cite{xiao2023efficient} and $\text{H}_2\text{O}$~\cite{zhang2023h}, typically employ a cache eviction policy that discards vectors less important for the generative inference process. However, such an approach, often rigid in nature, tends to overlook the potentially valuable information that might be contained within these deemed less important vectors. Our experimental findings reveal that this conventional approach does not perform optimally for multimodal instruction-following models.  

This insight highlights the need for a more nuanced cache management strategy to discern and retain valuable data, even if it might appear less important initially. Thus, we propose Cache Merging, which merges the surrounding less-important cached vectors into the select anchor vector. 
We determine the importance of each token in the instruction sequence and use the key/value vectors of tokens with high importance as anchor points, dividing the instruction sequence into multiple buckets. Each bucket corresponds to the neighborhood around an anchor point. 
Then, we store the averaged results of all the key/value vectors corresponding to each bucket in the KV cache.
By this merging policy, we reduce the number of key/value vectors in the KV cache and preserve the contextual information as much as possible. 

\subsection{One Sequence, Two Policies} 


In the context of multimodal instruction-following models, the model operates differently in the instruction encoding (perception, one-off feed-forward pass) and output generation (generation, iterative output process) stages. 
This divergence catalyzes implementing varied acceleration policies tailored to each specific stage of inference. We call this strategy ``\emph{One Sequence, Two Policies}''. 

\bfsection{Instruction Encoding}
During the instruction encoding stage, we adopt our \emph{importance-driven cache merging} policy, which unfolds in two primary steps: 1) selection of important key/value vectors with the layer-wise sum of attention scores as the metric and 2) division and merging of vectors based on selected anchors. 

Suppose the model has $L$ layers in total, each with $K$ heads. Let's consider the \textbf{\textit{j}}-th attention head in the \textbf{\textit{i}}-th layer, denoted as $\text{Attn}_{i,j}$. The input to $\text{Attn}_{i,j}$ is a sequence of $T$ tokens $\{x^{i,j}_{1}, \ldots, x^{i,j}_{T}\}$. For each token in the sequence, we first obtain the query, key, and value vectors using linear projection. For a specific token $x^{i,j}_{t}$, its query, key, and value vectors are denoted as $q^{i,j}_{t}$, $k^{i,j}_{t}$, and $v^{i,j}_{t}$, respectively. For the attention head $\text{Attn}_{i,j}$, we can derive a causal attention matrix $A$ of size $T \times T$, which is a lower triangular matrix. For simplicity, we omit the superscripts on the attention matrix here.
This matrix can be written as:
\begin{equation} \label{eq:attention}
\begin{aligned}
A_{m,n} = 
\begin{cases} 
\frac{\exp(\langle q^{i,j}_{m}, k^{i,j}_{n} \rangle)}{\sum_{m'<m}\exp(\langle q^{i,j}_{m}, k^{i,j}_{m'} \rangle)} & \text{if } m \geq n \\
0 & \text{if } m < n 
\end{cases}
\end{aligned}
\end{equation}
where $\langle q_{m}^{i,j}, k_{n}^{i,j} \rangle$ represents the dot product of $q_{m}^{i,j}$ and $k_{n}^{i,j}$.

Based on this attention matrix, we can calculate the importance score of each token. For the $n$-th token, we can define its importance value as obtained from the $\text{Attn}^{i,j}$ module, denoted by $I_n^{i,j} = \sum_{m} A_{m,n}^{i,j}$. Then, by averaging the importance values obtained from all attention heads in the same layer, the importance value of the $n$-th token in the $i$-th layer can be expressed as $I_n^{i} = \frac{1}{K} \sum_{j} \sum_{m} A_{m,n}^{i,j} $. Consequently, we obtain independent importance scores for each layer, implying that our pruning strategy is layer-wise.

Given a predefined retention ratio $\gamma$, we select the top $N_I = \gamma \times T$ tokens in $I_n$, with the indices $\{ t_k \mid k = 1, 2, \ldots, N_I \}
$ in ascending order. 
Taking them as anchor points, we can have $N_I$ buckets as 
\begin{equation} \label{eq:bucket}
\begin{aligned}
B_k = \begin{cases}
\{0, ..., \left\lfloor \frac{t_{1} + t_{2}}{2} \right\rfloor\}, &  k = 1\\
\{\left\lfloor \frac{t_{k-1} + t_{k}}{2} \right\rfloor+1, \ldots, \left\lfloor \frac{t_{k} + t_{k+1}}{2} \right\rfloor \}, &  1 < k < N_I \\
\{\left\lfloor \frac{t_{N_{I}-1} + t_{N_{I}}}{2} \right\rfloor, ..., T\}, &  k = N_I
\end{cases},
\end{aligned}
\end{equation}
where $\left\lfloor \cdot \right\rfloor$ denotes the floor function. 
The bucket division varies across different layers while remaining consistent for all attention heads within each layer. 
For each attention head, we average all its key/value vectors as $\text{KV}_k$ corresponding to each bucket $B_k$ and store them in the KV cache:
\begin{equation} \label{eq:merge}
\text{KV}_k = \frac{1}{|B_k|} \sum_{t \in B_k} kv_t .
\end{equation}
Notably, each attention head has its own unique $\text{KV}_k$; again, we omit the superscript for simplicity.

\bfsection{Output Generation}
The output generation stage functions iteratively, updating the KV cache at each iteration. It necessitates that the cache management policy be adaptable for continuous increment. 
The conventional methods, exemplified by $\text{H}_2\text{O}$~\cite{zhang2023h}, adhere to the policy in the instruction encoding stage that accumulates attention scores and prioritizes cached vectors with higher scores. However, this approach appears less effective in the context of new token generation. Compared to the already cached vectors, the newly generated tokens lack an inherent advantage under this scoring system.
As the generation progresses, the likelihood of newly generated tokens being retained in the cache diminishes incrementally, making this method less sensible for dynamically evolving caches. 

To address this issue, we introduce a \emph{fixed-point elimination} strategy, in light of the assumption that the initialized instruction guidance and closely related generation carry greater importance, whereas the earlier generated content holds less significance. 
Specifically, we employ a queue-based system where, upon generating a new token and adding its corresponding key/value vectors, we systematically remove earlier cached vectors from a fixed truncation point in the queue.
Consider a current KV cache denoted as $\{\text{KV}_k | k = 1, 2, \ldots, N_C \}$ where $N_C$ represents the current length of the cache. We define a fixed truncation location $N_{tl}$.
Upon adding a new token to the cache, we calculate the current token retention ratio and compare it with a predefined threshold. If the ratio is lower than the threshold, the token is retained. Otherwise, the token at the truncation location $N_{tl}$ is removed, leading to an updated cache represented as:
$\{\text{KV}_k | k = 1, 2, \ldots, N_{tl}-1, N_{tl}+1, \ldots, N_C+1\} $

Experimentally, selecting a fixed position $N_{tl}$ that is arbitrarily close to $N_I$ has often yielded impressive results. To approach this more systematically, this position can be regarded as a hyperparameter, which allows for tuning through model selection techniques.

\section{Experiments}

In this section, we conduct extensive experiments to illustrate the effectiveness and efficiency of \emph{Elastic Cache}. We use two mainstream LVLMs, LLaVA-1.5~\cite{liu2023visual, Liu2023ImprovedBW} and Qwen-VL~\cite{Bai2023QwenVLAF} as our backbone and adopt the \emph{Elastic Cache} on instruction-following chat generation datasets. As no benchmark exists on the efficient inference of the vision instruction tuning field, we design precise and general evaluation metrics to validate our method. We subsequently conduct experiments on inference speed, ablations on our method design, and real-world analysis.

\subsection{Experimental Settings.}

\paragrapha{Baselines. } To evaluate the effectiveness of our proposed caching mechanism for input instruction handling, we establish two state-of-the-art baseline methods for comparison: Heavy-Hitter Oracle~\cite{zhang2023h} and StreamingLLM~\cite{xiao2023efficient}, where we term as $\text{H}_2\text{O}$ and Local in our experiments, respectively. The techniques proposed by $\text{H}_2\text{O}$ and StreamingLLM can be summarized as Frequency Cache and Local Cache methods. Frequency Cache method records how often each key-value pair is accessed in response to input instructions. When the need arises to free up cache space to adhere to the pre-defined memory budget, this method selects the least frequently accessed key-value pairs for eviction. The Frequency Cache method continuously updates the frequency counts throughout the generation process. The Local Cache method employs a spatial heuristic for cache eviction. When the cache reaches capacity, the Local Cache method identifies and removes the most distant key-value pairs from the current prediction. 

\begin{figure*}[!t]
    \centering
    \includegraphics[width=\linewidth]{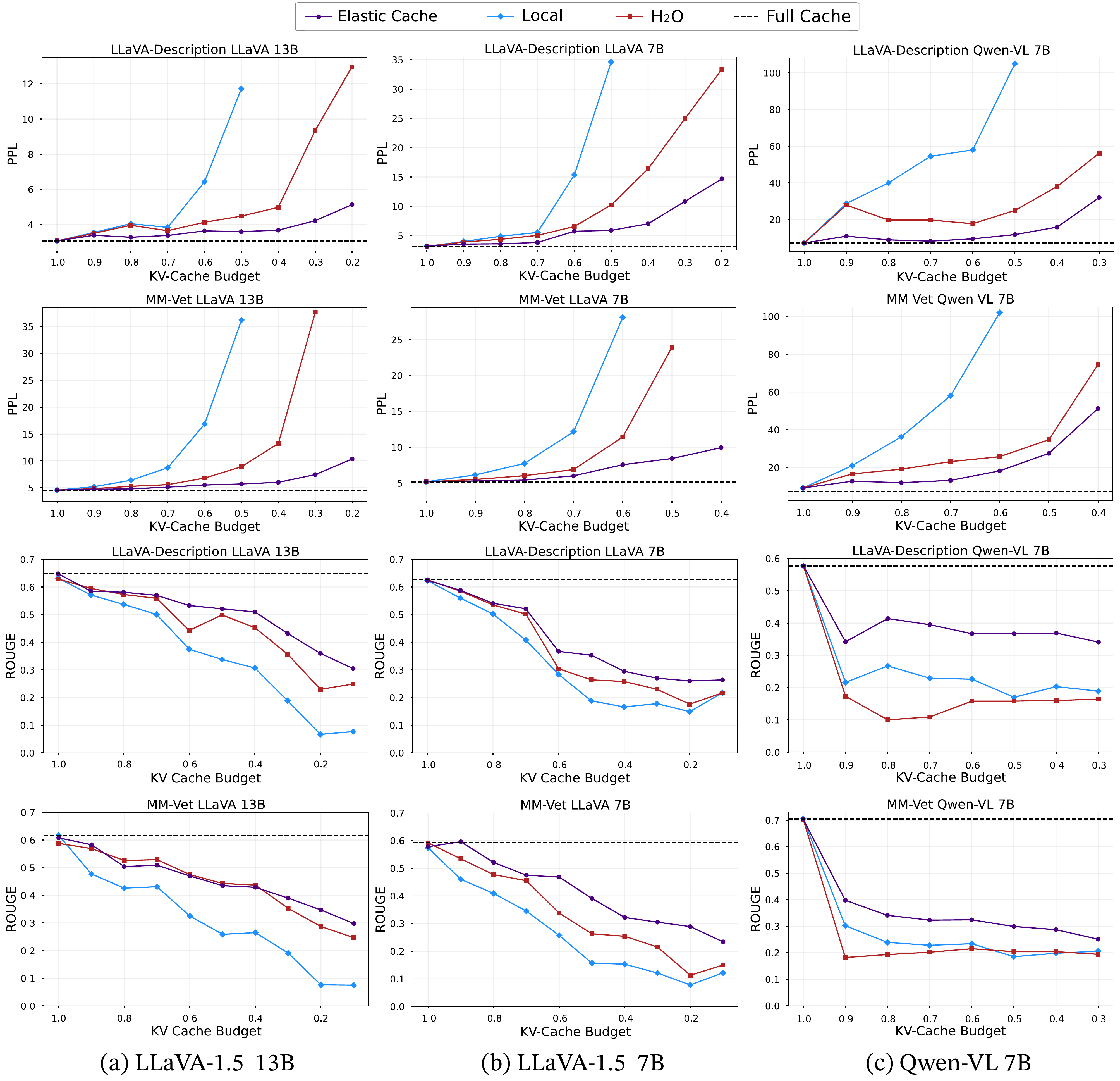}
    \caption{\textbf{Results on visual instruct-following tasks. }We evaluate \emph{Elastic Cache} together with baselines on PPL (lower better) and ROUGE (higher better) metrics. We conduct LLaVA-1.5 of different sizes (a),(b)  and Qwen-VL-7B(c) for visual tasks. Our \emph{Elastic Cache} outperforms baselines consistently.}
    \label{fig:main_result}
\end{figure*}

\paragrapha{Evaluation Metrics. }In the evaluation period of our method, we employ two metrics to assess performance rigorously: perplexity (PPL) and the ROUGE score~\cite{lin2004rouge}. PPL calculates the expositional value of the cross-entropy loss between the predicted next token and the ground truth. The ROUGE score measures the longest common subsequence (LCS) between the generated text and a set of reference texts. We use the F1-score to evaluate our methods. These metrics are chosen for their ability to capture distinct aspects of model performance, with perplexity reflecting the model's uncertainty in predicting the next token and the ROUGE score indicating the quality of text generation in terms of overlap with reference sequences. The detailed algorithms of the evaluation metrics are stated in Supplementary Materials. 

\paragrapha{Datasets. }\emph{Elastic Cache} is a training-free approach during the inference phase. To evaluate the text generation capabilities of the key-value (KV) cache strategy on long and high-quality text samples, we randomly integrate a subset of 100 detailed description instructions from the training set of LLaVA-1.5~\cite{Liu2023ImprovedBW} dataset, where we name as LLaVA-Description. Additionally, we also select the MM-Vet~\cite{Yu2023MMVetEL} dataset which covers a diverse range of tasks, enabling us to evaluate the model's multimodal understanding and generation performance comprehensively. Note that to prevent data leakage and ensure the integrity of our assessment, we redo the instruction tuning period of the LLaVA-1.5 models and exclude the aforementioned 100 detailed description instructions for the evaluation employing the KV cache strategy in our experiments.

\paragrapha{Implementation Details. }We mainly apply our proposed \emph{Elastic Cache} on three instruct-tuning VLMs. We choose LLaVA-7B/13B and Qwen-VL-7B as the visual instruct-tuning model. In our implementation of \emph{Elastic Cache}, we protect the first and the most recent token following~\cite{xiao2023efficient}. Following the analysis results in \cref{fig:distance}, we fixed the recent distance during the generation period at the length of 25 caches. It is noteworthy that for the reference texts required for the ROUGE evaluation, we use the fully-cached model to generate the reference text due to the ROUGE evaluation method. Notably, as the generation results are different during multiple experiments when the temperature is larger than zero, when the KV-Cache budget is set to 1.0, we use the ROUGE score of two-generation results, so the ROUGE score is smaller than 1.0. In other cases, we set the temperature to 0 to ensure the reproducibility of the results.

\subsection{Main Results}

\paragrapha{Results on Visual Instruct-Following. }Our experiments focus on visual instruction-following, utilizing a subset of the LLaVA dataset specifically curated for detailed description analysis.  This subset comprises 100 instances of image description tasks, with each instance including an instruction and a corresponding answer generated by GPT-4~\cite{OpenAI_GPT4_2023}. For our evaluation metrics, we employ Perplexity (PPL), where a lower score indicates better model performance, and the F1 variant of the ROUGE-L score, where a higher score reflects greater quality. We leverage the LLaVA-1.5/13B, LLaVA-1.5/7B, and Qwen-VL-7B as our backbone architectures. Compared to the baseline strategies, namely Local cache and $\text{H}_2\text{O}$ cache, our proposed \emph{Elastic Cache} demonstrates superior performance across both metrics over a spectrum of KV-Cache Budgets ranging from 1.0 to 0.2.  Notably, in \cref{fig:main_result}b, at a KV-Cache Budget of 0.5, \emph{Elastic Cache} surpasses the $\text{H}_2\text{O}$ cache by a margin of 4.34 in PPL and 0.089 in ROUGE-L F1 score on LLaVA-Description.  Compared to the Local cache, \emph{Elastic Cache} shows an improvement of 28.72 in PPL and 0.165 in ROUGE-L F1 score.  This enhancement in performance is likely attributable to the fact that the \emph{Elastic Cache}'s dynamic pruning strategy is more adept at retaining image-relevant knowledge, in contrast to the fixed strategy of most recent pruning, which can overlook critical visual information.

\setlength{\columnsep}{10pt}%
\begin{wraptable}{r}{8cm}
\setlength{\abovecaptionskip}{-0.8cm} 
  \centering
  \caption{\textbf{Win-rate comparison using GPT-4V API. } We primarily compare the win rate at different pruned ratios with existing work under GPT-4V. \emph{Elastic Cache} still obtains 38\% win-rate with a pruned ratio of 0.3, surpassing previous work by a large margin.}
    \adjustbox{width=\linewidth}{
    \begin{tabular}{cccc}
    \toprule
    ~Method~ & ~~~Budget=0.1~~~ & ~~~Budget=0.2~~~ & ~~~Budget=0.3~~~ \\
    \midrule
   Elastic Cache &  \textbf{47.54\%} &\textbf{46.63\%} & \textbf{37.56\%} \\
   $\text{H}_2\text{O}$~\cite{zhang2023h} & 38.55\% & 35.26\% & 30.26\% \\
   Local~\cite{xiao2023efficient} & 46.37\% & 35.29\% & 10.10\% \\
    \bottomrule 
    \end{tabular}
      \label{tab:win}}
\end{wraptable}

\paragrapha{Comparisons under \\ GPT-4V Evaluation. } We further conduct experiments incorporating GPT-4V~\cite{OpenAI_GPT4V} evaluation to demonstrate the effectiveness of our \emph{Elastic Cache} mechanism from a different perspective. In detail, we first collect 200 generations with the KV-Cache ratio 1.0 (the original generation procedure with full cache) for each method as our baseline. Then, we perform each method with a different pruned ratio to ask the same questions and forward the image-question pair and the corresponding answers to GPT-4V. For each method, GPT-4V is asked to determine whether the generated text is better or worse than the corresponding baseline. We follow the prompt proposed in FastGen~\cite{ge2023model} and modify it to customize it for image-question pairs. As shown \cref{tab:win}, our method illustrates superior robustness as the pruned ratio grows. $\text{H}_2\text{O}$ also demonstrates its robustness, but the lower win rate evaluated by GPT-4V indicates the unsatisfied capability compared with our method. On the contrary, Local exhibits competitive results at a low pruned ratio. However, it encounters a disastrous performance drop when the pruned ratio grows to 0.3, which further reveals the splendid ability of our method to accommodate both effectiveness and robustness.

\subsection{Inference Speed}

\begin{table}[t]
  \centering
  \caption{\textbf{Evaluations on inference latency and throughput. }We set the KV-Cache budget as 0.2 on the LLaVA-1.5/13B backbone. \emph{Elastic Cache} can lead to a maximum of 77.9\% actual speed up and avoid out-of-memory in limiting cases.}
    \adjustbox{width=\linewidth}{
    \begin{tabular}{ccccccc}
    \toprule
    ~~~Batch~~~ & ~~~Model~~~  & ~~~~~Token~~~~~ & \multicolumn{2}{c}{Latency (s)} & \multicolumn{2}{c}{Throughput (token/s)} \\
\cmidrule(lr){4-5}\cmidrule(lr){6-7} Size & Size & Length & ~~Elastic Cache~~ & ~~Full Cache~~ & ~~Elastic Cache~~ & ~~Full Cache~~ \\
    \midrule
    8     & 13B   & 1024+512   & 20.2$_{\cb{+(33.8\%)}}$  & 30.5  & 202.8$_{\cb{+(52.6\%)}}$ & 132.9 \\
    16    & 13B   & 624+256   & 11.8$_{\cb{+(34.1\%)}}$  & 17.9  & 347.1$_{\cb{+(51.7\%)}}$ & 228.8 \\
    \midrule
    16    & 7B    & 1024+512   & 17.2$_{\cb{+(43.8\%)}}$  & 30.6  & 476.3$_{\cb{+(77.9\%)}}$ & 267.7 \\
    48    & 7B    & 624+256   & 13.6$_{\cb{+N/A}}$  & OOM   & 903.5$_{\cb{+N/A}}$ & OOM \\
    \bottomrule
    \end{tabular}%
  \label{tab:speed}}
\end{table}

We evaluate the inference speed of our novel \emph{Elastic Cache} mechanism, implemented within the LLaVA-1.5/13B framework, to demonstrate its practical efficiency.  We conduct this evaluation under two distinct experimental configurations: the first with an input comprising 1024 prompt tokens followed by the generation of 512 tokens, and the second with a reduced input of 624 prompt tokens, and generate 256 tokens. The latter setup represents the minimal prompt length, including image patches and system prompts that can be processed. The inference tests are performed on a single NVIDIA A100 GPU, with batch size adjusted to maximize the available memory, thereby reflecting a realistic usage scenario optimized for both efficiency and throughput. Results are shown in \cref{tab:speed} Our results indicate that when pruning 80\% of the KV-Cache, the \emph{Elastic Cache} method can achieve an actual speed acceleration of up to 78\% compared to full cache baselines, while also reducing the memory footprint during inference.

\subsection{Ablation Experiments}
\label{sec:ablation}

\begin{table}[tb]
\caption{\small \textbf{Ablation studies on the components of \emph{Elastic Cache}.} We conduct four ablation studies on the key design of \emph{Elastic Cache}. The experiments are conducted on the LLaVA-1.5/13B backbone with the KV-Cache budget set as 0.5. We use PPL metric for the LLaVA detailed description dataset for evaluation. }
\centering
\subfloat[\footnotesize Ablation on \\discard position.]
{\makebox[0.245\linewidth][c]{
\adjustbox{width=0.245\linewidth}{
\tablestyle{8pt}{1.2}
\begin{tabular}{c|c}
\textbf{Discard Position} & \textbf{PPL} \\
\shline
 Most Recent & 3.93 \\
Frequency & 3.75 \\
\rowcolor{Gray} Fixed-point & \textbf{3.60} \\ 
\end{tabular}
}}}
\hfill
\subfloat[\footnotesize Ablation on \\merging strategy.]
{\makebox[0.245\linewidth][c]{
\adjustbox{width=0.245\linewidth}{
\tablestyle{8pt}{1.2}
\begin{tabular}{c|c}
\textbf{Merging Strategy} & \textbf{PPL} \\
\shline
 Clustering & 3.61 \\
Cache Eviction & 3.68 \\
\rowcolor{Gray} Cache Merging & \textbf{3.60} \\ 
\end{tabular}
\label{Tab:ablation:b}
}}
}
\hfill
\subfloat[\footnotesize Ablation on \\attention procedure.]
{\makebox[0.245\linewidth][c]{
\adjustbox{width=0.245\linewidth}{
\tablestyle{8pt}{1.2}
\begin{tabular}{c|c}
\textbf{Attn. Procedure} & \textbf{PPL} \\
\shline
Shared & 3.73 \\
Head-wise & 3.75 \\
\rowcolor{Gray} Layer-wise  & \textbf{3.60} \\ 
\end{tabular}
\label{Tab:ablation:c}
}}
}
\hfill
\subfloat[\footnotesize Ablation on \\importance metric.]
{\makebox[0.245\linewidth][c]{
\adjustbox{width=0.245\linewidth}{
\tablestyle{8pt}{1.2}
\begin{tabular}{c|c}
\textbf{Importance Metric} & \textbf{PPL} \\
\shline
Moving Average & 8.43 \\
Mean & 8.70 \\
\rowcolor{Gray} Sum & \textbf{3.60} \\ 
\end{tabular}
\label{Tab:ablation:d}
} 
}}
\label{tab:ablation}
\end{table}

We perform complete and detailed ablation experiments to evaluate the effects of each component in \emph{Elastic Cache} in \cref{tab:ablation}. 

\paragrapha{Discard Position. }The strategy during the output generation period contributes to maintaining the KV-Cache ratio with more generated caches. We explore how to discard older caches to improve the efficiency during output generations. Compared with leaving the most recent caches while discarding the farthest caches and continually taking the frequency policy during generation, we find that fixing a specific discard position can achieve better performance. 

\paragrapha{Merging Strategy. }Compared with simply evicting the chosen caches, we find that merging the unused caches with validated caches can achieve better performance. We also compare our recent merging method with the clustering algorithm, which clusters the unused cache to 10 cluster centers based on the key of caches. Though the clustering method involves extra cache cost, the overall performance is poorer than the simple recent merging method.

\paragrapha{Attention Procedure. }As we obtain the attention score across the layer and head dimensions, we are curious about whether applying the dynamic results across such dimensions contributes to better performance. We can conclude from the results that the layerwise strategy performs better than leaving an identical strategy across layers. However, including the dynamic nature in the multi-head dimension will lead to an accuracy drop. 

\begin{table}[t]
  \centering
    \caption{\textbf{Generations on image recognition question.} We fix the KV-Cache Budget as 0.5. Local and $\text{H}_2\text{O}$ cache pruning methods fail to generate rational results under such experimental settings while \emph{Elastic Cache} maintains the generation ability with a detailed and correct description of the image. }
  \begin{tabular}{@{}ll@{}}
    \toprule
    \hskip 0.1cm
    \begin{minipage}{0.4\linewidth}
      \includegraphics[width=\linewidth]{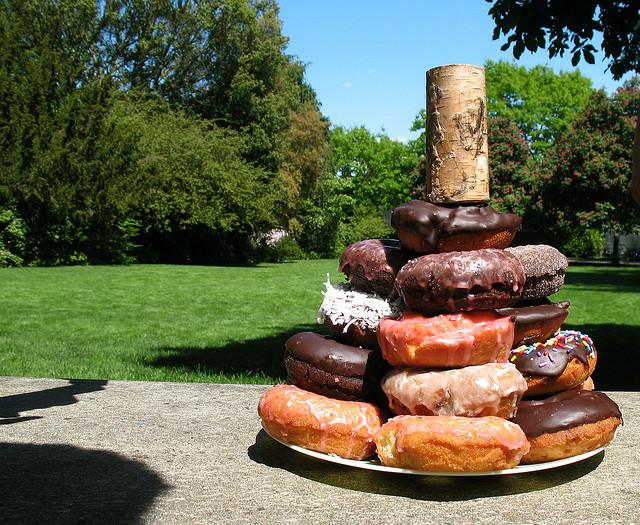}
    \end{minipage}
    \hfill
    \hskip 0.2cm
    \begin{minipage}{0.5\linewidth}
      \footnotesize{\textbf{User:} What's happening in the scene? \\
      \textbf{Local:} The image shows a plate of assorted doughnuts on a table, The doughnuts are arranged in a plate, with a total of 10 doughnuts on a table, showcasing a total of 10 doughnuts on a table ... \\
      \textbf{$\text{H}_2\text{O}$:} The image shows a park scene with a doughnut on a table.\\
      \textbf{Elastic Cache:} The image features \textbf{a large pile of assorted donuts}, including \textbf{glazed and chocolate} donuts, arranged in a visually appealing display. ... making it an \textbf{attractive sight} for anyone passing by.}
    \end{minipage} \\
    \bottomrule
  \end{tabular}
  \begin{tabular}{@{}ll@{}}
    \toprule
    \hskip 0.1cm
    \begin{minipage}{0.4\linewidth}
      \includegraphics[width=\linewidth]{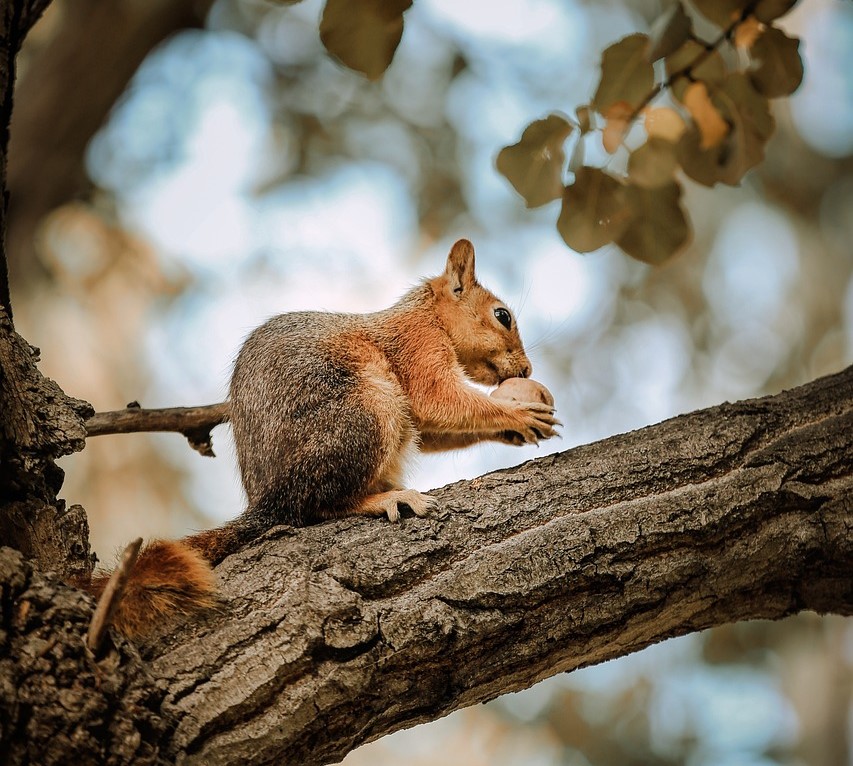}
    \end{minipage}
    \hfill
    \hskip 0.2cm
    \begin{minipage}{0.5\linewidth}
      \footnotesize{\textbf{User:} What's happening in the image? \\
      \textbf{Local:} The squirrel, a squirrel, a small gray squirrel, which appears to be a grey squirrel, is perched on a tree branch, sitting on a tree branch. \\
      \textbf{$\text{H}_2\text{O}$:} The squirrel is holding on to a tree branch.\\
      \textbf{Elastic Cache:} The squirrel is \textbf{sitting on a tree branch}, holding onto it while \textbf{eating}. The tree is surrounded by a forest setting, and the squirrel is \textbf{enjoying its meal}.}
    \end{minipage} \\
    \bottomrule
  \end{tabular}
  \label{tab:generation}
\end{table}

\paragrapha{Importance Metric. }The way of calculating the importance metric of each cache based on the attention score is a crucial problem to explore. We try the three most simple strategies to statistic the attention score. By using the moving average, mean, and sum value of the attention score, we observe that simply sum up the attention score outperforms another method to a large extent.

\subsection{Analysis}

\paragrapha{Generation.}  In the empirical analysis of our method's performance in real-world text generation scenarios, we rigorously evaluate the robustness of our caching strategy under constrained conditions in \cref{tab:generation}. We set the Key-Value (KV) Cache Budget to 0.5. Our method demonstrates remarkable resilience in the face of significant cache limitations. Notably, when 50\% of the KV cache is evicted, our approach continues attending to image details and producing coherent and rational outcomes. In contrast, as the cache budget decreases, the Local and $\text{H}_2\text{O}$ strategies suffer a conspicuous performance degradation. With the KV Cache budget set to 0.5, the Local strategy begins generating repetitive, looping text and fails to respond. Similarly, the $\text{H}_2\text{O}$ strategy yields significantly shorter responses, suggesting an inability to sustain longer, more intricate narratives when deprived of less frequently accessed cache content.

\setlength{\columnsep}{10pt}%
\begin{wrapfigure}{r}{8cm} 	
  \centering
  \vspace{-20pt}
    \includegraphics[width=\linewidth]{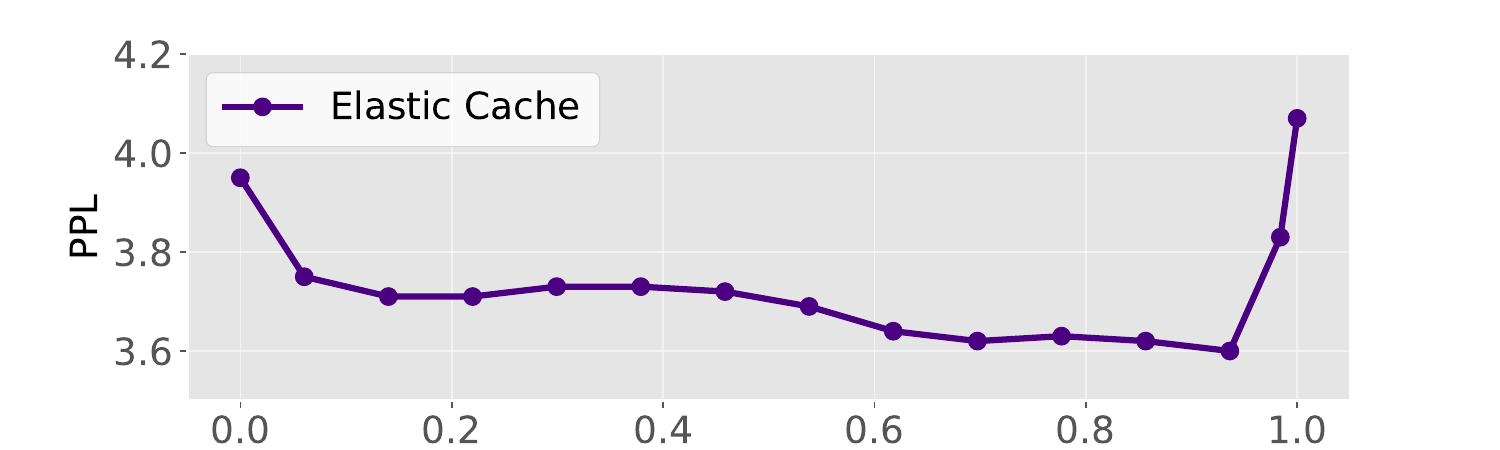}
    \caption{\textbf{Effects of the fixed-point elimination.} We observe that fixing the elimination at the middle of the KV-Cache of the instruction attention score leads to better performance.}
    \setlength\abovedisplayskip{-2cm}
    \label{fig:distance} 
\end{wrapfigure}

\paragrapha{Effects of the fixed-point elimination. }We explore the efficacy of a fixed-point elimination strategy within the caching mechanism, seeking to determine the optimal position for fixation to enhance performance.  Intrigued by the potential impact of different fixed points on the overall system effectiveness, we conduct a series of tests across various positions in \cref{fig:distance}.  The empirical evidence suggests that anchoring the fixed point within the middle section of the cache sequence yields superior results.  Guided by these findings, we strategically fix the position at the 25 most recent caches based on the experimental results. 

\section{Conclusion, Limitations and Societal Impact}

In this paper, we have proposed \emph{Elastic Cache}, an innovative framework designed to significantly enhance the efficiency of inference processes in widely utilized instruction-following models.  Our novel approach leverages the importance-driven cache merging strategy, which utilizes attention scores as a measure of importance to optimize cache utilization.  We further dissect the instruction-following paradigm into two distinct components: instruction encoding and output generation, applying the most effective strategy independently. The experimental results are compelling. \emph{Elastic Cache} not only surpasses existing baselines but also demonstrates robust generation capabilities coupled with remarkable speed improvements. We hope our work can open a new path for the following work to explore a better strategy and efficient inference of vision large models.

One limitation of our approach could be that the reliance on attention scores for cache optimization may not always align with the most computationally efficient caching strategy. A potential negative social impact is that the increased efficiency and speed of \emph{Elastic Cache} might accelerate the deployment of AI systems in surveillance applications.

\section*{Acknowledgements}
This work was supported in part by the National Key Research and Development Program of China under Grant 2023YFB280690, and in part by the National Natural Science Foundation of China under Grant 62321005, Grant 62336004, and Grant 62125603.

%
%
\bibliographystyle{splncs04}
\bibliography{main}

\begin{appendix}

\section{Implementation Details}

In this section, we will provide more details about our implementations of instantiating \emph{Elastic Cache}.

\subsection{Metrics}

In the evaluation phase of our experiments, we employed two distinct metrics to gauge the performance of the accelerated inference process. Firstly, Perplexity (PPL) was utilized as a measure of predictive accuracy, quantifying the model's ability to forecast the next token in a sequence given the ground-truth texts. This metric is particularly critical in assessing the immediate, token-level precision of the model, providing insight into its understanding of the language structure on a granular level. Secondly, we incorporated the ROUGE score, as defined by Lin~\cite{lin2004rouge}, to evaluate the overall coherence and fidelity of the generated texts in comparison with the ground-truth references. Together, these metrics furnish a robust framework for assessing the dual aspects of linguistic accuracy and contextual relevance in the generated text, thereby delivering a holistic picture of the model's performance post-acceleration.

In our study, the Perplexity (PPL) score serves as a critical metric for evaluating the language model's performance, and we adhere to the methodology delineated by Xiao et al. (2023)~\cite{xiao2023efficient} for its computation. The perplexity is essentially a measure of how well a probability distribution or probability model predicts a sample. To calculate it, we first determine the Cross-Entropy loss for each predicted word $\hat{w}_i$ in relation to the corresponding ground-truth word $w_i$, given the sequence of true preceding words $w_1w_2\cdots w_{i-1}$. This loss quantifies the discrepancy between the predicted probability distribution and the actual distribution of the words. Once we have computed the Cross-Entropy losses for all $N$ positions in the text, these losses are aggregated and the exponential of the average loss is taken to yield the overall PPL score for the entire text. Mathematically, the process of calculating perplexity can be encapsulated as follows:
\begin{equation}
\begin{aligned}
&\text{PPL}=e^{l} \\
&l=\frac{1}{N}\sum_{i=1}^{N}\text{Cross-Entropy}(\hat{w}_i|w_1w_2\cdots w_{i-1}, w_i)
\end{aligned}
\end{equation}
Here, $p(w_i | w_1w_2\cdots w_{i-1})$ represents the model's predicted probability for the ground-truth word $w_i$, given the sequence of preceding words. The negative logarithm of this probability is the Cross-Entropy loss for each word, and the average of these losses across the text is exponentiated to compute the PPL. This metric is particularly telling as it encapsulates both the model's fluency and its ability to predict subsequent words in a sequence, with lower scores indicating a model that is better at predicting the sample.

In the manuscript, we adopt the ROUGE score as a principal evaluation metric to quantify the similarity between sentences produced by the model and the target sentences.  Specifically, we utilize the ROUGE-L score, which is predicated on identifying the longest common subsequence (LCS) between the output of our model and the reference text.  It is important to note that the LCS is not required to be contiguous;  rather, it is a sequence that appears in both the generated text and the reference, albeit potentially with other intervening words.  We compute the ROUGE-L precision, which reflects the proportion of the LCS that is present in the model-generated output.  Concurrently, we assess the ROUGE-L recall, representing the proportion of the LCS found within the reference text. Subsequently, we synthesize these insights into a single metric by calculating the F1-score as follows:
\begin{equation}
\text{F1(ROUGE-L)}=\frac{2\cdot(\text{precision}\cdot\text{recall})}{(\text{precision}+\text{recall})}
\end{equation}

\subsection{Pseudo Code}

In order to elucidate our proposed \emph{Elastic Cache} mechanism, we have included a Pytorch-style pseudocode representation in the manuscript. This pseudocode, presented in \cref{alg:code}, serves to bridge the gap between conceptual understanding and practical implementation, providing readers with a clear, step-by-step guide to the algorithm's operational framework. It is crafted to mimic the syntactical and structural conventions familiar to users of the Pytorch library, thereby ensuring that the logic and flow of the \emph{Elastic Cache} are both accessible and immediately applicable to practitioners in the field. More details can be found in the \texttt{code} folder. By doing so, we aim to facilitate the reproducibility of our results and enable other researchers to seamlessly integrate or build upon our caching strategy within their own computational models. 

\begin{algorithm}
\caption{\emph{Elastic Cache} PyTorch-like Style Pseudocode. }\label{alg:handmim}
\begin{algorithmic}
\State \textbf{Define} 
\State Cache size $N$, Fixed position $P$, Ratio $r$, Sum of score $S$
\State \textbf{Input}
\State Past KV-Caches $kv$, Number of tokens $n$, Attention scores $attn$
\If{$kv$ is None}
    \State \textbf{Return} None
\EndIf
\State seq\_len = $kv$[0][0].size(2)
\State gen\_len = $attn$[0].size(-2)
\State $attn$ = $attn$.mean(dim=1)
\If{gen\_len $>$ 1}
    \State $S$[:, seq\_len] = $attn$.sum(dim=-1)
\EndIf
\State del\_num=int(seq\_len - $n$ * (1 - $r$))
\If{del\_num $\leq$ 0}
    \State \textbf{return} $kv$
\ElsIf{del\_num $\geq$ 1}
    \State $kv\_new$ = []
    \State fix\_idx=seq\_len - del\_num + $P$
    \For{$i$, $k$, $v$ in enumerate($kv$)}
        \State keep\_idx = where(argsort($S$[$i$,:seq\_len])$>$del\_num)
        \State throw\_idx = where(argsort($S$[$i$,:seq\_len])$\leq$del\_num)
        \State merge\_idx = Nearest(throw\_idx, keep\_idx)
        \State $k\_$throw=$k$.gather(dim=-2,index=throw\_idx)
        \State $v\_$throw=$v$.gather(dim=-2,index=throw\_idx)
        \State $k$=$k$.scatter\_reduce(dim=-2,index=merge\_idx)
        \State $v$=$v$.scatter\_reduce(dim=-2,index=merge\_idx)
        \State $k\_new$=$k$.gather(dim=-2,index=keep\_idx)
        \State $v\_new$=$v$.gather(dim=-2,index=keep\_idx)
        \State $kv\_new$.append([$k\_new$,$v\_new$])
    \EndFor
    \State \textbf{return} $kv\_new$
\Else
    \State $kv\_new$ = []
    \For{$i$, $k$, $v$ in enumerate($kv$)}
        \State $k\_new$=cat([$k$[:fix\_idx],$k$[fix\_idx+1:]], dim=-2)
        \State $v\_new$=cat([$v$[:fix\_idx],$v$[fix\_idx+1:]], dim=-2)
        \State $kv\_new$.append([$k\_new$,$v\_new$])
    \EndFor
    \State \textbf{return} $kv\_new$
\EndIf
\end{algorithmic}
\label{alg:code}
\end{algorithm}

\section{Chatting}

\begin{figure}[H]
    \centering
    \includegraphics[width=0.8\linewidth]{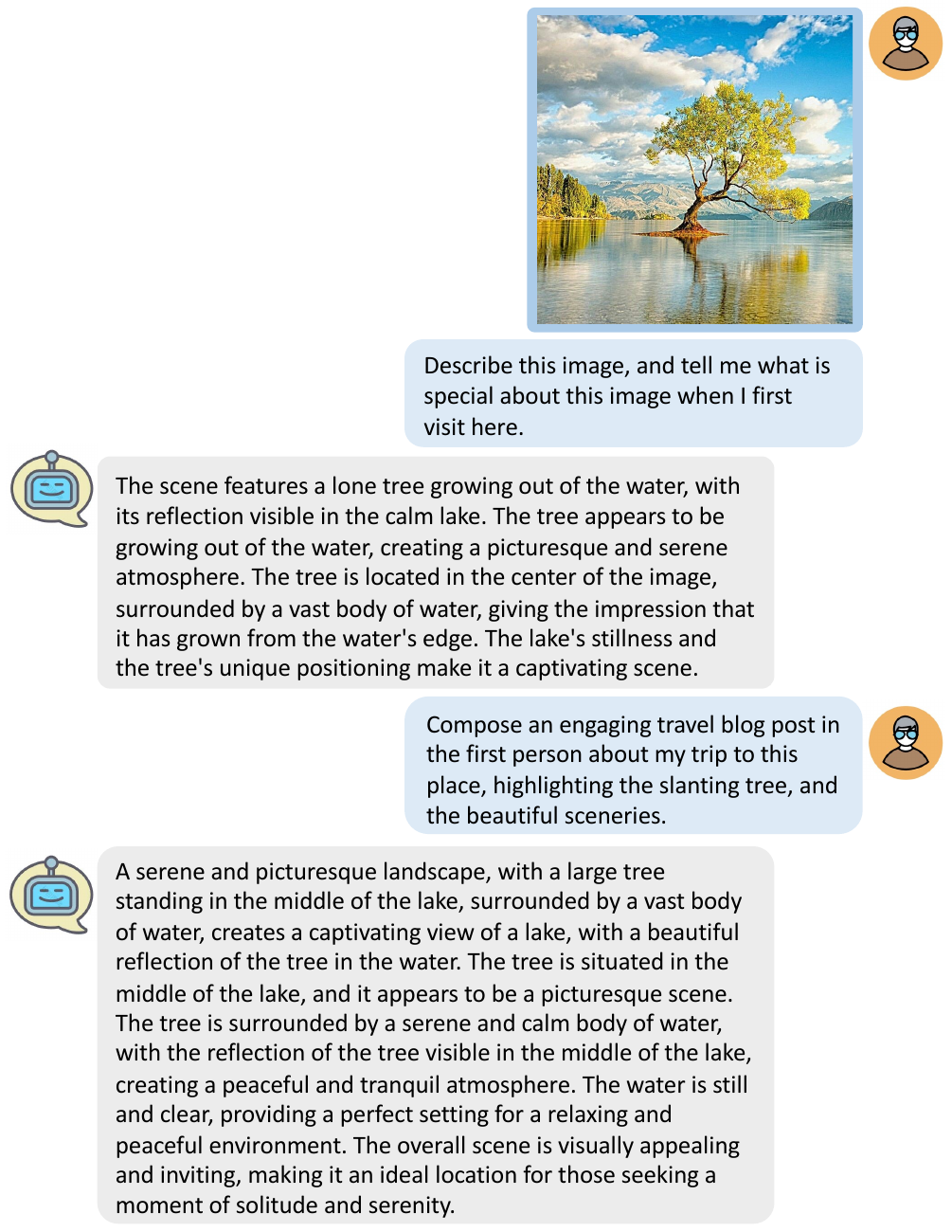}
    \caption{\textbf{Chat generation demo.} We set the KV-Cache Budget as 0.5 and generate a response under the dialogue setting. The LLM generates meaningful and abundant outputs. }
    \label{fig:chat}
\end{figure}

We showcase the capabilities of our \emph{Elastic Cache} framework through a practical demonstration of interactive chat generation, as depicted in Figure \ref{fig:chat}. The language model, powered by the \emph{Elastic Cache}, engages in a dialogue with users, generating responses that are not only contextually relevant but also rich in content and diversity.

\section{More Generation Results}

\begin{table}[t]
    \centering
    \caption{\emph{Elastic Cache} can correctly answer user questions and provide more detailed descriptions and explanations. As a comparison, the focus of the Local and H2O cache pruning methods is incorrect, and they may even be unable to answer questions. } 
    \includegraphics[width=5cm]{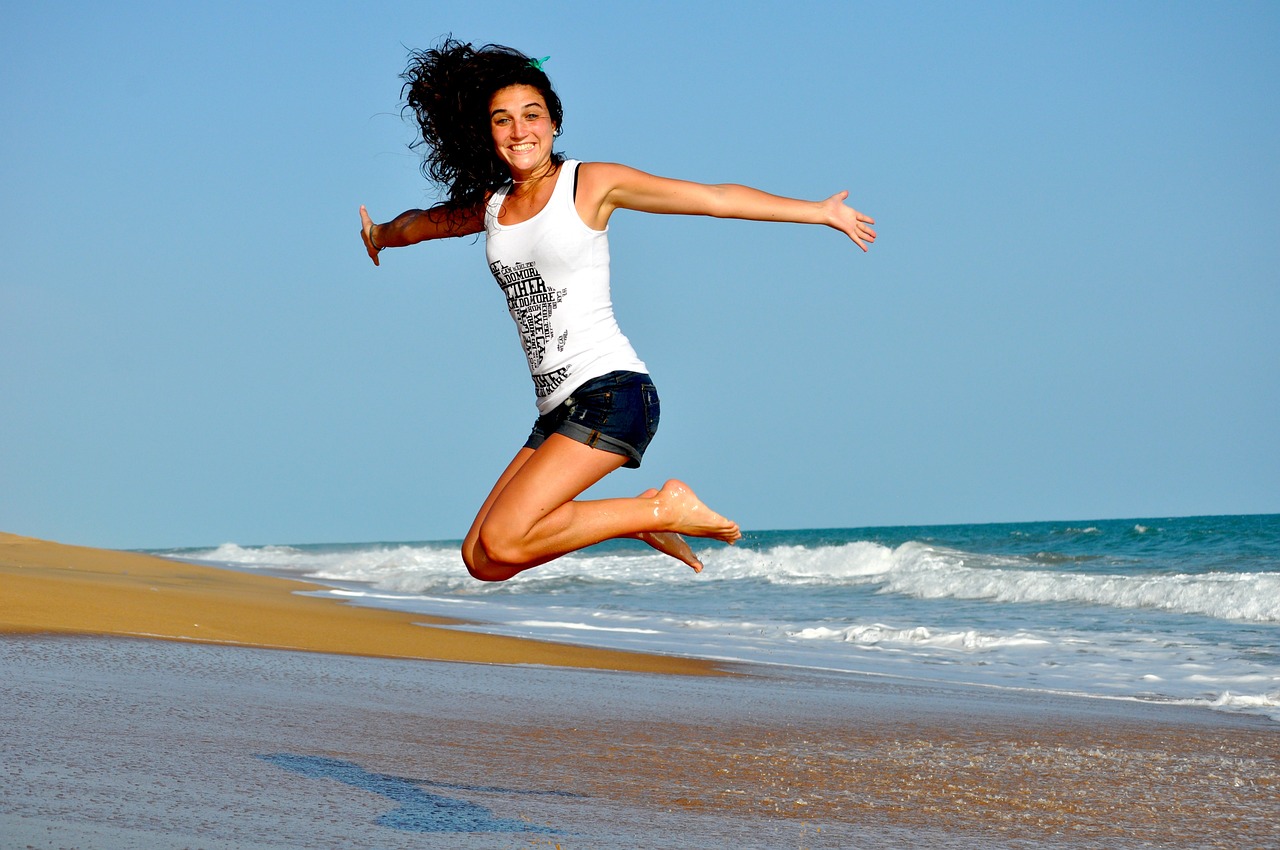} 
    \adjustbox{width=\linewidth}{
    \begin{tabular}{@{}ll@{}}
        \toprule
        User  &  Imagine the woman's mood in this image.               \\
        \midrule
        \multicolumn{2}{l}{\emph{KV-Cache Budget=0.5}} \\
        \midrule
        Local  & The woman is captured mid-air, enjoying the beach volleyball. \\
        \midrule
        H2O  & The woman is jumping up and down as she runs along the beach. \\
        \midrule
        \emph{Elastic} & The woman is wearing a white tank top and shorts, and she is jumping on the \\
        \emph{Cache} & beach. She is smiling and appears to be \textbf{having a great time}.\\
        \midrule \midrule
        \multicolumn{2}{l}{\emph{KV-Cache Budget=0.2}} \\
        \midrule
        Local & Thesrellsssssssssssssssst. \\
        \midrule
        H2O & The woman is smiling and appears happy and enjoying herself on the beach. \\
        \midrule
        \emph{Elastic} & The woman in the image is captured in the photo is likely  to be \textbf{happy and}\\
        \emph{Cache} & \textbf{excited}. She is smiling and \textbf{enjoying the moment}. She has a smile on her face,\\
              &  ... The woman's posture suggests that she is enjoying the moment and \textbf{having}\\
              & \textbf{a good time}. \\
        \bottomrule
    \end{tabular}
    }
    \label{tab:woman}
\end{table}

Our comparative analysis of cache strategies for image-text generation is meticulously documented in Tables \cref{tab:woman}, \cref{tab:home}, and \cref{tab:baseball}, where we explore the efficacy of these strategies under various Key-Value (KV) Cache budget constraints. The tables present a detailed examination of how each strategy performs when the available cache is limited, providing a comprehensive understanding of their capabilities and limitations.

\begin{table}[t]
    \centering
    \caption{\emph{Elastic Cache} is capable of maintaining focus on key elements in the image, such as bookshelves, sofas, and chairs, even when the budget is set to 0.2, and it provides reasonable answers. In contrast, the Local and H2O methods exhibit text repetition and may even fail to answer questions when the budget is increased to 0.5. } 
    \includegraphics[width=5cm]{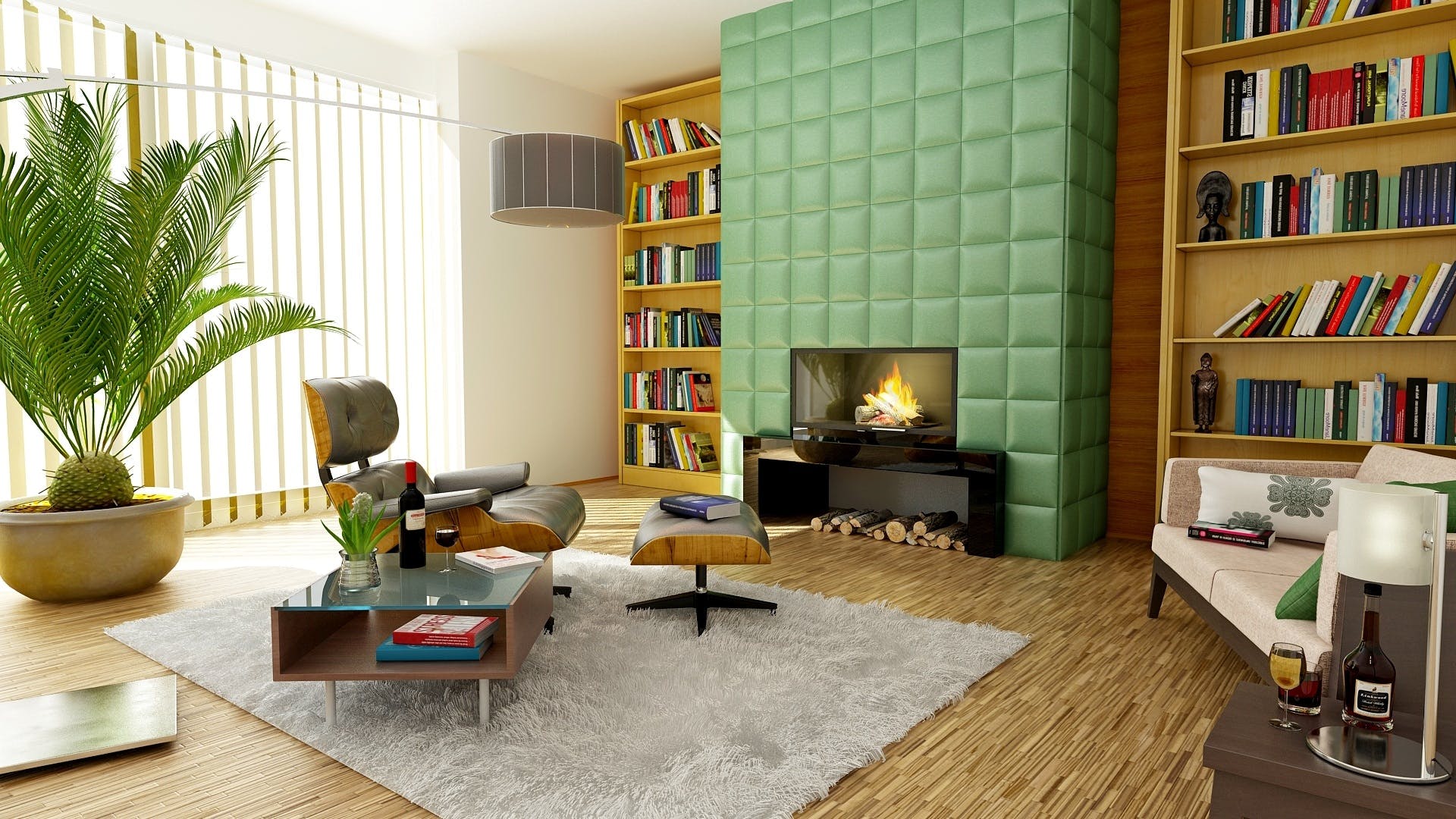} 
    \adjustbox{width=\linewidth}{
    \begin{tabular}{@{}ll@{}}
        \toprule
        User  &  Describe this image.               \\
        \midrule
        \multicolumn{2}{l}{\emph{KV-Cache Budget=0.5}} \\
        \midrule
        Local  & The living room is a cozy, modern and well-decorated living room with a fireplace.\\
              &  The living room is a modern and well-decorated living room with a fireplace. The \\
              & room features a fireplace with a fireplace. ...\\
        \midrule
        H2O & The room is filled with furniture, including a couch, a chair, and a coffee table. \\
              & ... The room has a green and white color scheme, with a green accent wall, and \\
              & a green accent wall. \\
        \midrule
        \emph{Elastic} & The living room is filled with \textbf{furniture and decorative items}, creating a cozy\\
        \emph{Cache} &  and inviting atmosphere. ... There is \textbf{a large bookcase filled with numerous} \\
              & \textbf{books}, showcasing a variety of books. The overall design of the room is modern \\
              & and inviting, with \textbf{a large window} allowing natural light to fill the space.\\
        \midrule \midrule
        \multicolumn{2}{l}{\emph{KV-Cache Budget=0.2}} \\
        \midrule
        Local & Therex. \\
        \midrule
        H2O & The image shows a living room with a large bookshelf, which has a large bookcase.  \\
        \midrule
        \emph{Elastic} & The image features a cozy living room with \textbf{a large bookcase filled with} \\
        \emph{Cache} &  \textbf{various books}, creating a warm and inviting atmosphere. The room is furnished\\
              &    with \textbf{a comfortable couch and a chair}. \textbf{A potted plant} is placed in the \\
              &  corner, adding a touch of greenery to the space. \\
        \bottomrule
    \end{tabular}
    }
    \label{tab:home}
\end{table}

\begin{table}[!t]
    \centering
    \caption{\emph{Elastic Cache} is able to identify the main focal points of an image with a lower budget, providing accurate responses regarding the relationships between individuals in the picture and making reasonable inferences about the context of a competition.} 
    \includegraphics[width=5cm]{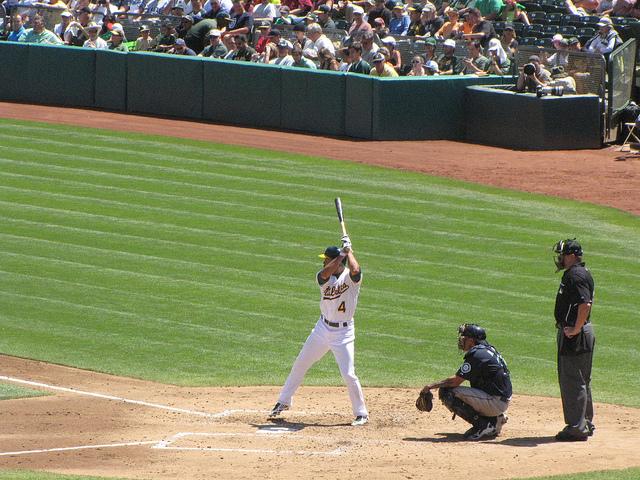}
    \adjustbox{width=\linewidth}{
    \scriptsize 
    \begin{tabular}{@{}ll@{}}
        \toprule
        User  &  Describe this image.               \\
        \midrule
        \multicolumn{2}{l}{\emph{KV-Cache Budget=0.5}} \\
        \midrule
        Local  & The image shows a baseball player in a white uniform, swinging a bat, ready\\
              & to hit the ball. The catcher is wearing a white uniform, and is standing in \\
              & the dirt, holding a bat, preparing to hit the ball. ... The catcher is wearing \\
              & a black shirt, holding a bat, getting ready to hit the ball. ... \\
        \midrule
        H2O  & The batter is holding a baseball bat, ready to swing at the pitch. The catcher \\
              & and umpire are behind him. \\
        \midrule
        \emph{Elastic} & The scene takes place on a baseball field, where a baseball game is in progress.\\
        \emph{Cache} &  The batter is \textbf{in the middle of } the action, holding a baseball bat and pre-\\
              & paring to swing at an incoming pitch. The catcher and umpire \textbf{are positioned}\\
              & \textbf{behind him}, ready to \textbf{catch the ball} if the batter misses. ... The scene \\
              & captures the intensity and focus of a professional baseball game.\\
        \midrule \midrule
        \multicolumn{2}{l}{\emph{KV-Cache Budget=0.2}} \\
        \midrule
        Local & The scene. \\
        \midrule
        H2O & The crowd of people are watching the game. \\
        \midrule
        \emph{Elastic} & The image shows a baseball game in progress, with a batter at home plate,\\
        \emph{Cache} &  holding a baseball bat, and getting ready to swing at a pitch. The scene cap-  \\
              & tivating moment in the game. The batter is \textbf{in the middle of the action,} \\
              & and the crowd is watching intently. The batter is ready to swing. The umpire\\
              & is \textbf{standing behind the catcher}, and the catcher is prepared to \textbf{catch the} \\
              & \textbf{ball}.The scene captures the intensity and excitement of a baseball game. \\
        \bottomrule
    \end{tabular}
    }
    \label{tab:baseball}
\end{table}

In the scenario depicted in \cref{tab:woman}, the challenge escalates with a nuanced question, \emph{The woman's mood}, posed to the model. Here, both the Distance and Frequency strategies buckle under the cache restrictions, unable to deliver accurate responses. Our \emph{Elastic Cache} strategy, however, stands out by not only correctly discerning the mood of the woman but also providing a nuanced description, even within the confines of a 0.2 budget, thereby underscoring the robustness of our method in dealing with complex interpretive tasks.

In the more complex scenes of \cref{tab:home} and \cref{tab:baseball}, where the model is tasked to \emph{Describe this image}, the limitations of the Distance and Frequency strategies become more pronounced, with both strategies succumbing to repetitive text generation. In contrast, the \emph{Elastic Cache} strategy maintains its composure and continues to deliver accurate depictions of the spatial relationships and interactions within the room and on the baseball field, respectively, even when operating on a meager 0.2 budget. This demonstrates the strategy's adaptability and its potential for handling scenarios with a higher degree of complexity.

In summary, our experimental findings across various budgets and scenarios firmly establish the \emph{Elastic Cache} strategy as the superior choice for image-text generation in resource-limited conditions. Remarkably, it consistently yields reasonable and coherent text outputs, even at the extremely constrained budget of 0.2, thereby attesting to its high applicability and robustness for practical use. This investigation contributes novel insights and methodologies to the domain of image-text generation and offers a valuable reference for addressing challenges in resource-constrained environments.

\end{appendix}

\end{document}